# Automatic essay scoring: leveraging Jaccard coefficient and Cosine similarity with n-gram variation in vector space model approach

Andharini Dwi Cahyani[1], Moh. Wildan Fathoni[1], Fika Hastarita Rachman[1], Ari Basuki[2], Salman Amin[3], Bain Khusnul Khotimah[1]
[1]Department of Informatics Engineering, Faculty of Engineering, Universitas Trunojoyo Madura, Bangkalan, Indonesia
[2]Department of Industrial Engineering, Faculty of Engineering, Universitas Trunojoyo Madura, Bangkalan, Indonesia
[3]School of Media and Communication Studies, Minhaj University, Lahore, Pakistan

**ABSTRACT**

Automated essay scoring (AES) is a vital area of research aiming to provide efficient and accurate assessment tools for evaluating written content. This study investigates the effectiveness of two popular similarity metrics, Jaccard coefficient, and Cosine similarity, within the context of vector space models (VSM) employing unigram, bigram, and trigram representations. The data used in this research was obtained from the formative essay of the citizenship education subject in a junior high school. Each essay undergoes preprocessing to extract features using n-gram models, followed by vectorization to transform text data into numerical representations. Then, similarity scores are computed between essays using both Jaccard coefficient and Cosine similarity. The performance of the system is evaluated by analyzing the root mean square error (RMSE), which measures the difference between the scores given by human graders and those generated by the system. The result shows that the Cosine similarity outperformed the Jaccard coefficient. In terms of n-gram, unigrams have lower RMSE compared to bigrams and trigrams.

*Corresponding Author:*

Andharini Dwi Cahyani
Department of Informatics Engineering, Faculty of Engineering, Universitas Trunojoyo Madura
Bangkalan, Indonesia
Email: andharini.cahyani@trunojoyo.ac.id

## 1. INTRODUCTION

Conventional essay tests give pupils a chance to demonstrate their intellectual diversity by presenting original ideas and points of view. Proficiency in written communication is an essential competency in both academic and professional contexts [1]. Students can show that they can analyse difficult situations, formulate arguments, and suggest answers by writing essays. It evaluates their ability to answer problems, which is crucial in a lot of real-world situations. Essays assess a student's ability for persuasive and lucid idea expression [2]. Organisation, coherence, and clarity are all essential communication skills in a variety of academic and professional settings.

Because essay evaluations are subjective, teachers are able to take into account students' individual writing preferences, viewpoints, and inventiveness. This adaptability is useful for assessing a range of answers. Nevertheless, despite its benefits, manual essay test assessment poses serious difficulties for teachers [3]. Researchers are increasingly addressing biases and ethical issues in manual essay scoring, working to detect and eliminate unfairness. Automated essay scoring (AES) systems have emerged as a game-changing solution, offering a more efficient and objective assessment method [4]–[6]. The limitations of manual grading may be addressed with an AES system, which provides a quick and efficient way to assess students' written work. The subject of AES research is vigorous, examining a range of methods, strategies, and technological improvements for automatically grading and scoring essays [7]. There is a rising interest in making AES models more comprehensible and explainable. Understanding how models result in specific ratings is critical for increasing trust in automated systems, particularly in educational settings. AES research has benefited greatly from the application of natural language processing (NLP) and text mining techniques [8]. Essays can yield valuable information through the use of sentiment analysis, syntactic analysis, documents resemblance, and semantic analysis [9], [10]. These methods aid in comprehending the text's sentiment, content, similarity, and organisation.

Andersen *et al*. [11] proposed the AES framework for assessing Danish writing proficiency in terms of text structure, sentence form, and modifier usage. They explored NLP and machine learning approaches to solve the problem. The research methodology employed in this study mainly based on the analytical framework suggested by Kabel *et al*. [12] for analysing early writing. Within this architecture, each text goes through two phases: statistical Rasch modelling for scoring, and annotation by a human expert following a predefined classification scheme. They carried out experiments to compare and assess the two approaches. Their results show that the scores generated by the automatic technique and the ones established by human experts have a strong correlation and are statistically significant.

Süzen *et al*. [13] explored automatic grading of short answers and providing insightful feedback using a dataset from the University of North Texas's Introductory Computer Science course. They applied the vector space model (VSM) to measure the similarity between student responses and model answers based on commonly used terms. They analyzed the correlation between these similarities and the scorers' assigned grades. Additionally, they used the k-means clustering method to group student responses, assigning the same score and identical feedback to answers within each cluster. The clusters represented groups of students with similar performance, determined by comparing terms in student answers with those in the model answer. According to the research cited above, text mining techniques can be used to develop objective scoring standards based on quantifiable linguistic elements collected from essays. Essays and other textual data can be analysed and their content understood using text mining approaches by utilizing sophisticated NLP algorithms [14], [15]. This makes it possible for the system to comprehend, parse, and extract relevant elements from the essays, including sentence structure, semantic significance, coherence, and language usage [16], [17]. This study contributes to the field of AES by evaluating Cosine similarity and Jaccard coefficient metrics with various n-gram variations (unigrams, bigrams, trigrams) to improve essay scoring accuracy. It explores semantic similarity between student essays and model answers while analyzing how n-grams capture both individual words and word sequences for better context. The research identifies which metric and n-gram combination best correlates with human scoring.

In relation to AES, the VSM has been extensively researched. Essay documents using VSM are modelled into vectors in a high-dimensional space [18], [19]. Every dimension is related with a distinct term (word or n-gram), and the vector's values reveals the significance or occurrence of those phrases in the essay documents (refer to Figure 1).

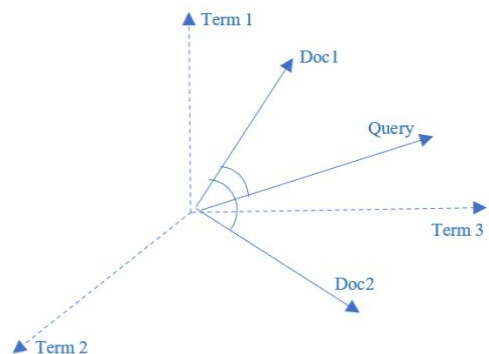

Figure 1. Illustration of document similarity in a vector space model

When creating these vectors, we usually utilize vectorization techniques, such as bag-of-words (BoW) [20], [21] and term frequency-inverse document frequency (TF-IDF) [22], [23]. Essays can be more nuancedly analysed and compared by managing NLP techniques that capture the semantic relationships between words and phrases that transform them into vector representations using the vector matrix [19], [24]–[26]. The VSM is commonly used for transforming textual data into numerical forms. VSM allows for the comparison of articles based on the similarity of their vector representations [27]–[29]. Because of its simplicity, VSM may be used with a wide range of machine learning models, providing flexibility in the selection of algorithms [30], [31].

## 2. RESEARCH METHOD

VSM is a mathematical model used to describe documents in vector form [32], [33]. Each document is represented as a vector in the same dimensional space with the number of dimensions equivalent to the number of words [34], [35]. Our research focuses on the usage of VSM for AES, which allows for the extraction of meaningful features from essays, capturing the relationships between words and their importance in the context of scoring. The TF-IDF weighting helps in emphasizing words that are important in a specific essay while downweighting common terms [36]–[38]. Once the essays are represented as TF-IDF vectors, Cosine similarity is commonly used to measure the similarity between essays. Cosine similarity calculates the cosine of the angle between two vectors and ranges from -1 (completely dissimilar) to 1 (completely similar) [39], [40]. In some cases, Jaccard similarity may be used, especially if the focus is on binary presence/absence of terms rather than their frequency [41], [42]. Figure 2 shows the process that is carried out in our study.

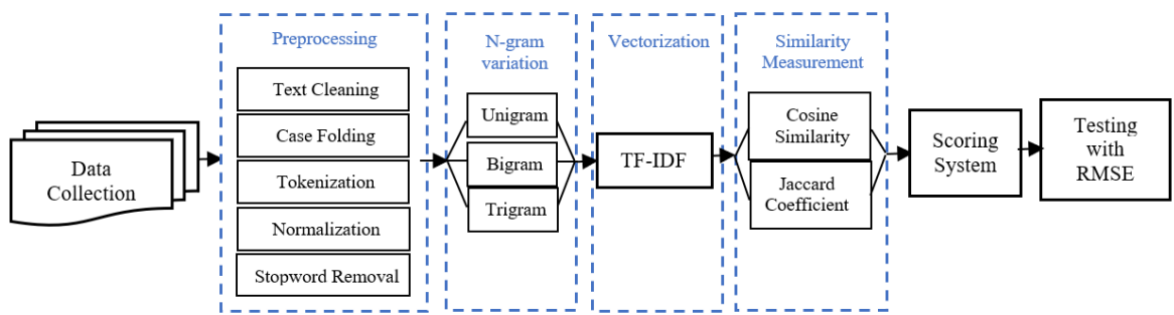

Figure 2. Research method

### 2.1. Data collection

The study collected data from 30 eighth-grade students at Junior High School Asa Cendekia Sidoarjo, specifically from the citizenship education subject. For formative assessment, the teacher gave the students 5 essay questions to assess their understanding of the material, resulting in a total of 150 essay responses in the dataset. The test was conducted on paper, and the students' answers were then converted into Excel format for further analysis.

### 2.2. Preprocessing

Text preprocessing is a crucial step in text mining, transforming raw text into an analyzable format. Its main goals are to enhance data quality, reduce noise, and extract meaningful information [43]–[45]. Here are the specific steps used in this study.
− Text cleaning: the goal of text cleaning is to improve the data quality by removing irrelevant or noise elements [46]. The process of text cleaning might vary based on the nature of the data and the needs of the research [47]. In our study, we used a text cleaning step to get rid of multiple spaces, punctuation marks, and non-alphabetic characters. This aids in keeping the dataset more structured.
− Case folding: this step involves transforming all characters, both uppercase and lowercase letters are converted to lowercase [48]. The goal of case folding is to standardize the text data, making it easier to compare, search, and analyze [49]. This process is important in NLP tasks where case sensitivity is usually not required and could lead to unnecessary complexity.
− Tokenization: in this phase, the document is divided into smaller parts called tokens. Tokenization is mainly used to break up continuous text into readily processed discrete parts, which are words [50], [51]. The analysis of NLP is built on tokens [52].

- Normalization: this step includes replacing slang and typo words with formal words taken from a dictionary. We also expand contractions words and abbreviations to ensure consistency in text representation. The goal of this step is to transform text data into a more consistent format [53].
- Stopword removal: stop words are words that often do not influence much to the meaning of a text [54], [55]. Stopwords are common in most documents, and their high frequency can lead to unnecessary computation time during text analysis [56]. Removing stop words can reduce the computational difficulty and focus on more meaningful content [57], and speed up the text processing.

To sum up, text preprocessing in our study refers to prepare raw text data for analysis. The steps in text preprocessing are depend on the goals of the text mining task and the nature of the dataset [58]–[60]. Implementing these stages effectively helps to prepare the text data for subsequent analysis, lower the computational complexity, and make it more suitable for machine learning models and other text mining techniques.

### 2.3. N-gram variation

NLP tasks normally use n-grams to identify and comprehend patterns in textual document [61]. Such NLP application that uses n-grams are language modelling, machine translation, document similarity comparation, and text generation [62], [63]. N-grams are essential to essay scoring, which aims to measure the similarity between students' answer with the model answer. N-gram models aid in capturing the links between words in a sequence and the surrounding information [64]. The text is tokenized in pre-processing processes to separate it into individual words or tokens before producing n-grams [65], [66].

In our study, n-grams are employed as features to represent the content and structure of the text. The VSM is a mathematical approach that transforms each essay document as a vector in a high-dimensional space, where each dimension representing a unique feature [67]. Each unique n-gram becomes a feature in the VSM. The presence or absence of these features is then used to represent the essay [68], [69]. The n-grams formation processes the conversion of essay document into a high-dimensional vector, where each dimension corresponds to the occurrence or absence of a specific n-gram.

N-gram feature extraction captures word sequences from a text document to describe its linguistic structure. Figure 3 illustrates how the feature space expands exponentially with an increase in "n" in n-grams, which can result in larger dimensionality and more computing complexity. Which n-gram size to use depends on the particular task at hand as well as the properties of the data being examined. Various n-gram sizes may be appropriate for different jobs.

```
Unigram: | This | is | a | sentence |         4 token
Bigram:  | This is | is a | a sentence |      3 token
Trigram: | This is a | is a sentence |        2 token
```

Figure 3. N-gram tokenization model

Unigrams are often used for basic text analysis tasks and initial feature extraction. Bigrams capture some level of local context and are useful for tasks like sentiment analysis. Trigrams provide a bit more context and are employed in tasks where understanding the relationships between three consecutive words is important.
- Unigram (1-gram): a unigram is a single word, representing the simplest form of n-gram. In unigram feature extraction, each word in the document is treated as a separate feature.
- Bigram (2-gram): a bigram is a sequence of two adjacent words. In bigram feature extraction, pairs of consecutive words are considered as features.
- Trigram (3-gram): a trigram is a sequence of three adjacent words. In trigram feature extraction, triplets of consecutive words are treated as features.

### 2.4. Vectorization

Vectorization in NLP converts text data into numerical representations for machine learning algorithms. A common method is TF-IDF, which assigns weights to words based on their frequency in a document (TF) and rarity across a corpus (IDF) [70]. This method reflects the significance of terms not just within a document, but across an entire collection [71]. TF-IDF applies to unigrams, bigrams, and trigrams, with context and meaning increasing with higher n-grams. Unigrams focus on individual words, bigrams on word pairs, and trigrams on more complex patterns.

TF: quantifies how often a term appears in a document [72]. It is calculated by dividing the number of times a term occurs by the document's total number of terms, as shown in (1), where: $f(t, A)$ represents the number of times term ttt appears in document A, $\sum f(w, A)$ denotes the total count of all terms in document A. IDF: assesses a term's importance across a document collection [73]. It's calculated by taking the logarithm of the ratio of the total number of documents to the number of documents containing the term, as shown in (2), where: | A | represents the total number of documents in the corpus. $DF(t)$ denotes the number of documents that contain the term t.

TF-IDF calculation: is obtained by multiplying the TF and the IDF for each term in the essay [74]. The formula to calculate TF-IDF is written as in (3). The resulting TF-IDF scores create a weighted representation of terms in the essay, emphasizing terms that are both frequent in the document and rare in the overall corpus.

$$TF(t, A) = \frac{f(t,A)}{\sum f(w,A)} \qquad (1)$$

$$IDF(t, A) = \log \frac{|A|}{DF(t)} \qquad (2)$$

$$TF - IDF(t, d, D) = TF(t, D) \times IDF(t, D) \qquad (3)$$

### 2.5. Similarity metric

Before scoring student essays, a set of reference answers is created to represent exemplary responses. A similarity metric is then applied to compare student essays with these references, generating a score that reflects their alignment. The choice of metric depends on the task and data characteristics, as text similarity metrics are widely used in NLP for various applications [75]. These metrics are interpretable, scalable for large datasets, and require careful selection based on task-specific needs [76]–[78]. Jaccard and Cosine similarity are both text similarity measurement method but serve different purposes [39], [41]. Jaccard similarity measures the proportion of shared terms between two documents relative to their total unique terms, making it useful for assessing text overlap and identifying near-duplicate responses. In contrast, Cosine similarity evaluates the angle between document vector representations in a high-dimensional space, making it effective for capturing semantic similarity even when documents have different lengths.

### 2.6. Cosine similarity

Cosine similarity is a similarity metric between two vectors in a dimensional space, that measures angle between the document vector and the query vector [79]. Each vector represents the document being compared and a word in the query. The score range of Cosine similarity varies continuously between 0 and 1 [34], where 0 indicates that the two documents are entirely dissimilar, while 1 signifies that they are perfectly aligned in direction, regardless of their length differences. The Cosine similarity formula between two vectors can be written as in (4), where $d_j$ = vector of dj documents, q = vector of query documents, $\sum_{i=1}^{t} W_{ij}$ = total of the weights of word i in document j, and $\sum_{i=1}^{t} W_{iq}$ = total of the weights of words i in q.

$$\text{Sim}(d_j, q) = \frac{d_j \cdot q}{|d_j| \cdot |q|} = \frac{\sum_{i=1}^{t} W_{iq} \cdot W_{ij}}{\sqrt{\sum_{i=1}^{t}(W_{iq})^2 \cdot \sum_{i=1}^{t}(W_{ij})^2}} \qquad (4)$$

### 2.7. Jaccard similarity

Jaccard similarity is one of similarity metric method that can be applied to various text data representations, making it suitable for tasks where the presence or absence of terms is crucial [80]. Jaccard provides a straightforward measure of similarity based on set operations, making it interpretable and easy to understand. TF-IDF takes into account not only the frequency of terms but also their importance in the context of the entire corpus. This allows Jaccard similarity to capture meaningful term overlaps.

The Jaccard similarity coefficient is then calculated based on the TF-IDF vectors of two essays [81]. Jaccard score ranges between 0 and 1 continuously [80]; 0 means no shared terms between the two documents, while 1 means both documents contain the exact same terms. The Jaccard similarity between essays A and B is given in (5). In the context of TF-IDF, the "terms in common" refer to the set of terms that have non-zero TF-IDF values in both essays.

$$Jaccard(A, B) = \frac{|TF-IDF \text{ Terms in Common between } A \text{ and } B|}{|Total \text{ Distinct } TF-IDF \text{ Terms in } A \text{ and } B|} \qquad (5)$$

## 2.8. Essay scoring system

The scoring system in this AES is based on a similarity metric that compare a student's essay and model answers obtained from the teacher [6]. The chosen similarity metric (e.g., Cosine similarity and Jaccard similarity) is applied to compare the vector representation of the student's essay with each teacher's essay, which results a similarity score for each question. The individual similarity scores are multiplied by the weight of each question, and then aggregated to obtain an overall score for the student's essay.

## 2.9. Testing with root mean square error

Root mean square error (RMSE), mean squared error (MSE), and mean absolute error (MAE) are all metrics used to evaluate the performance of a predictive model [82], [83]. The choice of these metrics depends on the problem characteristic. In this study, we employ RMSE to evaluate the proposed AES model simply because RMSE is more sensitive to large errors than MSE and MAE. RMSE penalizes larger errors more heavily due to the squaring operation. This sensitivity can be advantageous where large errors are considered more critical and have a significant impact on the overall performance of the model [84].

RMSE is consistent with the standard deviation of the target variable [85]. It also allows for a direct comparison with the standard deviation, providing a sense of scale for the errors. This makes it easier to interpret the error compared to the variability of the data. The RMSE is the final metric that quantifies the average magnitude of the errors made by the model in predicting the scores. A lower RMSE shows better performance, as it signifies that the predictions are closer to the actual scores. Conversely, a higher RMSE suggests larger discrepancies between predicted and actual scores. The RMSE formula is as in (6), where Yt = score from teacher for each student, Ut = aggregation score from system, and n = total student.

$$RMSE = \sqrt{\frac{1}{n}\sum_{t=1}^{n}(Y_t - U_t)^2} \qquad (6)$$

## 3. RESULTS AND DISCUSSION
### 3.1. Results

This study used descriptive statistics and a one-way repeated measures ANOVA to test the null hypothesis that there is no statistically significant difference between the mean scores assigned by the AES VSM system and human teachers. Table 1 presents a comparison of these mean scores. In the context of evaluating scoring methods such as AES VSM with Cosine similarity, AES VSM with Jaccard coefficient, and human grading, a one-way ANOVA helps to assess whether the observed differences in their average scores are due to true differences in the methods or merely a result of random variation. Using a one-way ANOVA is crucial in experimental analysis because it provides a rigorous method for testing whether the means of different groups differ significantly, while controlling for variability and reducing the risk of errors. This allows researchers to make confident, data-driven decisions about the effectiveness or reliability of different methods.

Table 2 presents the results of a one-way repeated measures ANOVA comparing AES with Cosine similarity to human grading. The F-value of 6.48 indicates that the differences in scores between the two methods are notable compared to the variability within each group. With a significance value (p =0.025) below the threshold of 0.05, it is clear that the difference between AES with Cosine similarity and human grading is statistically significant.

Similarly, Table 3 highlights the results of a one-way repeated measures ANOVA comparing AES with Jaccard coefficient to human grading. The test shows a statistically significant difference (p =0.031), meaning the two methods produce distinct scoring patterns. The large effect size ($\eta^2$ =0.535) and low Wilks's lambda ($\Lambda$ =0.236) further underscore that the grading method strongly influences the scores. These results demonstrate that scores from AES with Jaccard coefficient differ significantly from those assigned by human grading. In this study, we implement the VSM method for AES system and investigate the usage of Cosine similarity and Jaccard similarity for unigram, bigram, and trigram. We compare the students' answers and model answer to get the similarity scores. Those scores are then evaluated with the score from the teacher, yielding the RMSE score. The RMSE score from our experiment is shown in Table 1. Lower RMSE values indicate better performance, as they indicate smaller errors between predicted and actual scores.

The lowest RMSE score across all testing scenarios in Table 4 is highlighted. For Cosine similarity, the minimum RMSE is 2.04, achieved with trigram. In contrast, for Jaccard coefficient, the lowest RMSE is 1.72, obtained using unigram. This indicates that the Cosine similarity performs better in this testing scenario, as it achieves a lower RMSE compared to Jaccard coefficient. Additionally, the performance seems influenced by the feature representation method (unigram, trigram), with trigram being more effective for Cosine similarity and Jaccard coefficient.

Table 1. Descriptive statistics

| Source | Means | Standard deviation | Coefficient of variation |
|---|---|---|---|
| AES VSM with Cosine similarity | 79 | 3.46 | 4.385 |
| AES VSM with Jaccard coefficient | 81 | 3.74 | 4.619 |
| Human grading | 78 | 2.45 | 3.140 |

Table 2. One-way repeated measures ANOVA (AES Cosine similarity vs human grading)

| Source | F | Wilks's Λ | Sig | η2 |
|---|---|---|---|---|
| Grading methods (AES with Cosine similarity vs human grading) | 6.128 | 0.312 | 0.025 | 0.572 |
| Error df | 8.70 | | | |

Table 3. One-way repeated measures ANOVA (AES Jaccard similarity vs human grading)

| Source | F | Wilks's Λ | Sig | η2 |
|---|---|---|---|---|
| Grading methods (AES with Jaccard coefficient vs human grading) | 7.128 | 0.236 | 0.031 | 0.535 |
| Error df | 9.00 | | | |

Table 4. Comparison of RMSE value for Cosine similarity and Jaccard similarity

| | RMSE cosine | | | RMSE Jaccard | | |
|---|---|---|---|---|---|---|
| | Unigram | Bigram | Trigram | Unigram | Bigram | Trigram |
| Question 1 | 2.33 | 2.06 | 2.04 | 2.72 | 2.91 | 2.3 |
| Question 2 | 2.89 | 6.09 | 8.32 | 3.4 | 5.89 | 7.95 |
| Question 3 | 3.2 | 4.56 | 5.31 | 3.62 | 4.92 | 5.94 |
| Question 4 | 5.09 | 3.1 | 4.05 | 2.73 | 4.35 | 5.46 |
| Question 5 | 2.38 | 3.79 | 6.76 | 3.38 | 6.33 | 8.43 |

Overall, based on Figure 4, Cosine similarity tends to perform better than Jaccard similarity, as indicated by the generally lower RMSE values across all n-gram representations and questions. This suggests that, based on the provided data, Cosine similarity might be providing a better fit to the actual scores. From our analysis, the Cosine similarity considers the frequency of tokens (words, n-grams) in the essays. It considers both the presence and frequency of words in the vectors. However, Jaccard similarity considers only the presence or absence of tokens, without accounting for their frequency. Using our dataset, it seems that the frequency of specific words or n-grams is crucial for scoring essays. Therefore, Cosine similarity may better capture these nuances, leading to lower RMSE values.

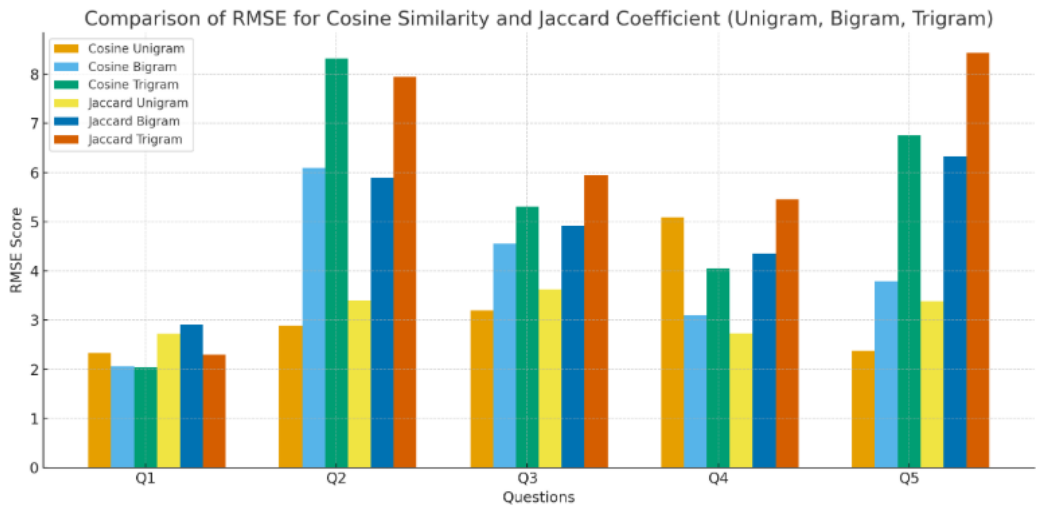

Figure 4. Comparison of RMSE in unigram, bigram, and trigram

## 3.2. Discussion

Our experiment shows that Cosine similarity performs better and has higher similarity to human grading compared to Jaccard coefficient. This result of our study is in line with that reported by Wahyuningsih *et al.* [86]. Their research showed that the Cosine similarity has similar performance with dice coefficient method and is better than Jaccard coefficient methods. Alobed *et al.* [41] also reported that the Cosine similarity has the lowest error compared with the Jaccard and Euclidean similarity in their automated Arabic essay scoring (AAES) application. Madatov and Sattarova [87] conducted an experiment to get highest performance of similarity metric using Cosine similarity, Jaccard similarity, and the combination of them. Their experiment shows that Cosine similarity outperformed the other two methods.

In general, Cosine similarity considers the magnitude of the vectors representing the data points in a high-dimensional space [88]. This means that even if the data is sparse and contains many zero values, Cosine similarity can still capture the similarity in direction between non-zero values, which is essential in high-dimensional spaces. Cosine similarity normalizes the vectors before computing the similarity, which mitigates the effect of varying magnitudes between data points. This normalization ensures that the similarity measure is not biased by the overall magnitude of the vectors, making it suitable for sparse data [89]. Moreover, in sparse data, where most of the values are zero (e.g., in text data represented as BoW or TF-IDF vectors), Jaccard similarity may not capture the similarity well because it only considers the presence or absence of non-zero values [90]. Cosine similarity, on the other hand, focuses on the angles between vectors and is less affected by the sparsity of the data. Since Cosine similarity focuses on the angles between vectors rather than the specific elements, it can handle high-dimensional, sparse data more effectively than Jaccard similarity.

Our experiment in Figure 4 shows that the impact of n-gram size varies across questions in Cosine similarity. For some questions (question 4), increasing the n-gram size leads to an increase in RMSE, indicates that greater n-gram size results less accurate in capturing the desired similarities between texts. Similarly, Citawan *et al.* [91] reported that their research in AES using latent semantic analysis (LSA) shows unigram have higher accuracy compared to bigram and trigram. Their research implied that variations of n-grams size show positive correlation in AES system. Combining neighbouring words into bigrams or trigrams captures more complicated text patterns and sentences. Compared to unigrams, bigrams and trigrams have more information, which could increase the model's complexity. Bigrams and trigrams often lead to feature spaces with higher dimensions, which in turn result in a greater level of sparsity in the representation [92]. The presence of sparsity might pose difficulties in the modelling process and may necessitate a larger amount of data to achieve good generalisation. If the model has difficulty capturing significant patterns in the data, the higher dimensionality can lead to increasing RMSE values.

Yazdani *et al.* proposed that unigrams create a lower-dimensional feature space compared to bigrams and trigrams, which helps reduce the risk of sparsity and overfitting [93]. Likewise, Li *et al.* indicated that a model with fewer dimensions is more likely to generalize effectively to previously unseen data, resulting in lower RMSE values [94]. In most essay documents, unigrams usually be seen more frequently rather than bigrams or trigrams. The form of unigrams produced more abundant tokens, therefore may offer a denser representation. Higher frequency may result to more stable and reliable representations, contributing to lower RMSE. Moreover, it is essential to note that the effectiveness of each similarity metric may vary subject to the specific characteristics of the NLP task and the nature of the dataset. Therefore, further experimentation and evaluation may be significant to explore the best similarity metric for the AES task.

### 3.3. Future research

Future research on AES could explore several promising directions. First, incorporating more advanced NLP techniques, such as transformer-based models like bidirectional encoder representations from transformers (BERT) or generative pre-trained transformer (GPT), could improve the system's ability to capture complex linguistic and semantic patterns. BERT, being a transformer-based model pre-trained on vast amounts of text data, excels at understanding context and generating rich, contextualized word and sentence embeddings. BERT works as a feature extractor, where the pre-trained model processes essay text to produce high-dimensional vectors that capture deep semantic meaning, coherence, and even subtle nuances that simple n-gram or TF-IDF representations miss. These sophisticated embeddings can then be fed into a separate regression or classification model to predict essay scores.

Second, expanding the datasets to include essays from diverse subjects and languages would enhance the system's generalizability and robustness. Additionally, integrating explainable AI methods can provide transparency into the scoring process, helping educators trust and adopt AES systems more widely. Finally, research can also focus on optimizing computational efficiency to ensure scalability for real-time applications in large educational settings. These advancements will help AES systems become more reliable, equitable, and accessible.

## 3.4. Implication

The adoption of AES has significant implications for educational practices and policy development. By providing an efficient and objective method for evaluating written assessments, AES systems have the potential to address long-standing challenges associated with manual grading, such as bias, inconsistency, and high workloads for educators [95]. This efficiency allows teachers to allocate more time to personalized instruction and mentoring, thus enhancing the overall quality of education. Additionally, AES fosters scalability in assessment practices, enabling institutions to evaluate large volumes of student essays in a timely manner without compromising fairness or accuracy.

The implementation of AES systems requires careful attention to ethical considerations, especially in terms of transparency, data privacy, and the risk of excessive reliance on automated tools [96]. For instance, transparency is crucial to ensure that students and educators understand how AES systems generate scores. Without clear explanations of the algorithms and criteria used, these systems could face skepticism or mistrust from stakeholders [97]. Moreover, there is a risk that over-reliance on AES could marginalize the role of educators, reducing their involvement in assessing student learning and providing valuable feedback [98]. For example, while AES can quickly score a large number of essays, it might struggle to recognize creative or nuanced responses that require human judgment. Policymakers and institutions must ensure that AES systems are deployed as supportive tools that enhance, rather than replace, the professional expertise of teachers [99]. Ultimately, AES represents a transformative tool in modern education, with the potential to enhance the objectivity and efficiency of assessments while supporting equitable educational opportunities. Ongoing research and collaboration between educators, technologists, and policymakers will be crucial in realizing its full potential while mitigating associated risks.

## 4. CONCLUSION

This study reveals valuable insight in the domain of AES by investigating the presentation of Jaccard coefficient and Cosine similarity metrics using the framework of VSM with n-gram variations. This research validates the preprocessing techniques and TF-IDF vectorization to get the document features by using a dataset from formative essays in citizenship education at the junior high school level. The comparison of Jaccard coefficient and Cosine similarity demonstrates that the latter surpasses the former in reviewing semantic similarity between documents. Moreover, the n-gram variations analysis discovers that unigrams lead to lower RMSE values compared to bigrams and trigrams, suggesting their ability in catching the main textual features. These findings highlight the consequence of selecting right similarity metrics and n-grams representations to lower the RMSE score of AES systems. Further research could study other factors influencing AES performance and investigate techniques for refining computational efficiency without compromising the performance. Ultimately, advancements in AES methodologies have the potential to revolutionize educational assessment practices, offering educators and stakeholders trustworthy tools for evaluating written content successfully.

## ACKNOWLEDGMENTS

The authors gratefully acknowledge the support provided by the Department of Informatics Engineering, Universitas Trunojoyo Madura, which facilitated the completion of this research.

## FUNDING INFORMATION

The authors declare that no external funding was received for conducting this study.

## AUTHOR CONTRIBUTIONS STATEMENT

This journal uses the Contributor Roles Taxonomy (CRediT) to recognize individual author contributions, reduce authorship disputes, and facilitate collaboration.

| Name of Author | C | M | So | Va | Fo | I | R | D | O | E | Vi | Su | P | Fu |
|---|---|---|---|---|---|---|---|---|---|---|---|---|---|---|
| Andharini Dwi Cahyani | ✓ | ✓ |  | ✓ | ✓ |  |  |  | ✓ | ✓ | ✓ | ✓ | ✓ |  |
| Moh. Wildan Fathoni |  | ✓ | ✓ |  |  | ✓ | ✓ |  | ✓ |  | ✓ |  |  |  |
| Fika Hastarita Rachman |  |  | ✓ | ✓ |  | ✓ |  | ✓ |  | ✓ |  |  |  |  |
| Ari Basuki | ✓ | ✓ |  |  | ✓ |  |  |  |  | ✓ |  |  |  | ✓ |
| Salman Amin | ✓ |  |  |  | ✓ |  |  |  | ✓ |  | ✓ |  |  |  |
| Bain Khusnul Khotimah |  | ✓ | ✓ |  |  | ✓ |  | ✓ | ✓ | ✓ |  |  | ✓ |  |

| | | | | | | | |
|---|---|---|---|---|---|---|---|
| C | : | **C**onceptualization | I | : | **I**nvestigation | Vi | : **Vi**sualization |
| M | : | **M**ethodology | R | : | **R**esources | Su | : **Su**pervision |
| So | : | **So**ftware | D | : | **D**ata Curation | P | : **P**roject administration |
| Va | : | **Va**lidation | O | : | Writing - **O**riginal Draft | Fu | : **Fu**nding acquisition |
| Fo | : | **Fo**rmal analysis | E | : | Writing - Review & **E**diting | | |

**CONFLICT OF INTEREST STATEMENT**

The authors state that there are no known competing financial interests or personal relationships that could have influenced the work reported in this paper.

**INFORMED CONSENT**

The authors affirm that informed consent was obtained from all individual participants included in this study, and written permission was secured prior to their inclusion.

**DATA AVAILABILITY**

The data that support the findings of this study are available from the corresponding author, [ADC], upon reasonable request.

## BIOGRAPHIES OF AUTHORS

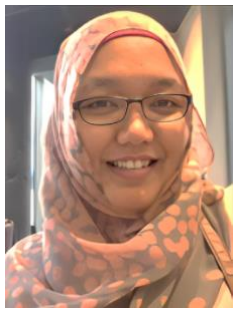

**Andharini Dwi Cahyani** obtained his bachelor's degree in Informatics Engineering from the Sepuluh Nopember Institute of Technology (ITS) in 2003 and completed her master's degree in the same institute in 2010. She later obtained her Ph.D. in Computer Science from Newcastle University, UK. She currently serves as an associate professor in the Department of Informatics Engineering at Universitas Trunojoyo Madura. Alongside her teaching responsibilities, she is actively involved in research and community engagement initiatives. Her primary research areas include software engineering, natural language processing, digital tourism, learning analytics, and decision support systems. She can be contacted at email: andharini.cahyani@trunojoyo.ac.id.

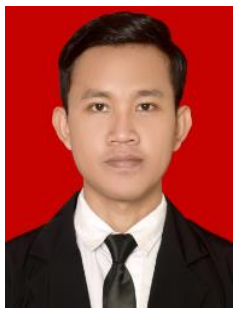

**Moh. Wildan Fathoni** obtained his Bachelor of Science (B.S.) degree in Informatics Engineering from Universitas Trunojoyo Madura. His academic and research interests are primarily focused on the fields of natural language processing, text mining, and automatic essay scoring. His work aims to contribute to advancements in both theoretical and applied aspects of these emerging technologies, particularly in enhancing the capabilities of language-based algorithms and systems. He can be contacted at email: wildanfathoni299@gmail.com.

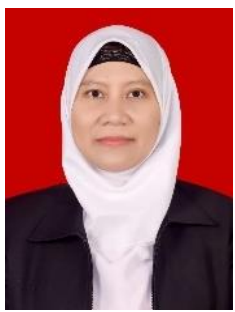

**Fika Hastarita Rachman** received the B.S. degree from the Department of Electrical Engineering, Universitas Brawijaya Malang, Indonesia, in 2005, and the M.S. degree from Department of Electrical Engineering and Information Technology, Universitas Gadjah Mada (UGM), Yogyakarta, Indonesia, in 2011, and she received the Ph.D. degree from the Department of Informatics, Institut Teknologi Sepuluh Nopember Surabaya (ITS). She is currently a lecturer at University of Trunojoyo Madura. Her research interests include data science, natural language processing, and text mining. She can be contacted at email: fika.rachman@trunojoyo.ac.id.

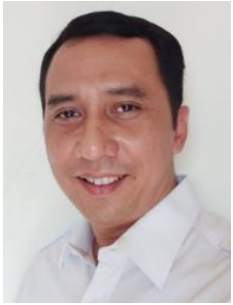

**Ari Basuki** obtained his bachelor's degree in Industrial Engineering from the Sepuluh Nopember Institute of Technology (ITS) in 2002 and completed his master's degree in Mechanical Engineering at Universitas Brawijaya in 2009. He currently serves as a lecturer in the Industrial Engineering program at Universitas Trunojoyo Madura. Alongside his teaching responsibilities, he is actively involved in research, community engagement initiatives, and university management activities. His primary research areas include industrial management, innovation management, and performance measurement. He can be contacted at email: aribasuki@trunojoyo.ac.id.

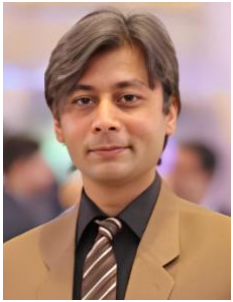

**Salman Amin** brings over nine years of experience in higher education and two years in the news media industry, establishing himself as a passionate and innovative professional in media and communication. With a Ph.D. in Mass Communication, his research delves into the influence of media on economic policy and societal transformation. In addition to his academic endeavours, he actively supports various research journals by offering editorial and technical expertise, reflecting his unwavering dedication to fostering growth in education and communication. He can be contacted at email: salmanamin.masscom@mul.edu.pk.

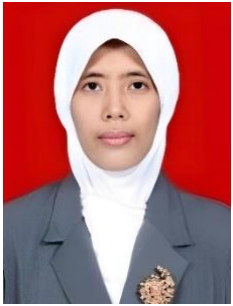

**Bain Khusnul Khotimah** is an associate professor in computer science with specialized expertise in data mining methodologies, focusing on machine learning and computational deep learning approaches. Her work encompasses practical applications within communities, including optimizing production processes through the adoption of appropriate technology, advancing product branding strategies, and developing information systems for measurable animal feed. Her recent research has concentrated on the computational analysis of animal data to identify superior breeds, supported by various research grants. She can be contacted at email: bain@trunojoyo.ac.id.